\documentclass[10pt,twocolumn,letterpaper]{article}

\usepackage{cvpr}
\usepackage{times}
\usepackage{epsfig}
\usepackage{graphicx}
\usepackage{amsmath}
\usepackage{amssymb}

\usepackage{url}
\usepackage{booktabs}
\usepackage{microtype}

\usepackage[pagebackref=true,breaklinks=true,letterpaper=true,colorlinks,bookmarks=false]{hyperref}

\cvprfinalcopy 

\def\cvprPaperID{451} 

\begin{document}

\title{Xception: Deep Learning with Depthwise Separable Convolutions}

\author{
  Fran\c{c}ois Chollet \\
  Google, Inc. \\
  {\tt\small fchollet@google.com} \\
}

\maketitle

\begin{abstract}

We present an interpretation of Inception modules in convolutional neural networks as being an intermediate step in-between regular convolution and the \textit{depthwise separable convolution} operation (a depthwise convolution followed by a pointwise convolution). In this light, a depthwise separable convolution can be understood as an Inception module with a maximally large number of towers. This observation leads us to propose a novel deep convolutional neural network architecture inspired by Inception, where Inception modules have been replaced with depthwise separable convolutions. We show that this architecture, dubbed Xception, slightly outperforms Inception V3 on the ImageNet dataset (which Inception V3 was designed for), and significantly outperforms Inception V3 on a larger image classification dataset comprising 350 million images and 17,000 classes. Since the Xception architecture has the same number of parameters as Inception V3, the performance gains are not due to increased capacity but rather to a more efficient use of model parameters.

\end{abstract}

\section{Introduction}

Convolutional neural networks have emerged as the master algorithm in computer vision in recent years, and developing recipes for designing them has been a subject of considerable attention. The history of convolutional neural network design started with LeNet-style models \cite{lecun1995learning}, which were simple stacks of convolutions for feature extraction and max-pooling operations for spatial sub-sampling.
In 2012, these ideas were refined into the AlexNet architecture \cite{krizhevsky2012imagenet},
where convolution operations were being repeated multiple times in-between max-pooling operations,
allowing the network to learn richer features at every spatial scale.
What followed was a trend to make this style of network increasingly deeper,
mostly driven by the yearly ILSVRC competition; first with Zeiler and Fergus in 2013 \cite{zeiler2014visualizing}
and then with the VGG architecture in 2014 \cite{simonyan2014very}.

At this point a new style of network emerged, the Inception architecture, introduced by Szegedy et al. in 2014 \cite{szegedy2015going} as GoogLeNet (Inception V1), later refined as Inception V2 \cite{ioffe2015batch}, Inception V3 \cite{szegedy2015rethinking}, and most recently Inception-ResNet \cite{inceptionResnet}. Inception itself was inspired by the earlier Network-In-Network architecture \cite{lin2013network}. Since its first introduction, Inception has been one of the best performing family of models on the ImageNet dataset \cite{russakovsky2014imagenet}, as well as internal datasets in use at Google, in particular JFT \cite{hinton15}.

The fundamental building block of Inception-style models is the Inception module, of which several different versions exist. In figure \ref{inception_module} we show the canonical form of an Inception module, as found in the Inception V3 architecture.
An Inception model can be understood as a stack of such modules. This is a departure from
earlier VGG-style networks which were stacks of simple convolution layers.

While Inception modules are conceptually similar to convolutions (they are convolutional feature extractors),
they empirically appear to be capable of learning richer representations with less parameters.
How do they work, and how do they differ from regular convolutions? What design strategies come after Inception?

\subsection{The Inception hypothesis}

A convolution layer attempts to learn filters in a 3D space, with 2 spatial dimensions (width and height) and a channel dimension; thus a single convolution kernel is tasked with simultaneously mapping cross-channel correlations and spatial correlations.

\begin{figure}[!ht]
  \caption{A canonical Inception module (Inception V3).}
  \label{inception_module}
  \centering
    \includegraphics[width=0.45\textwidth]{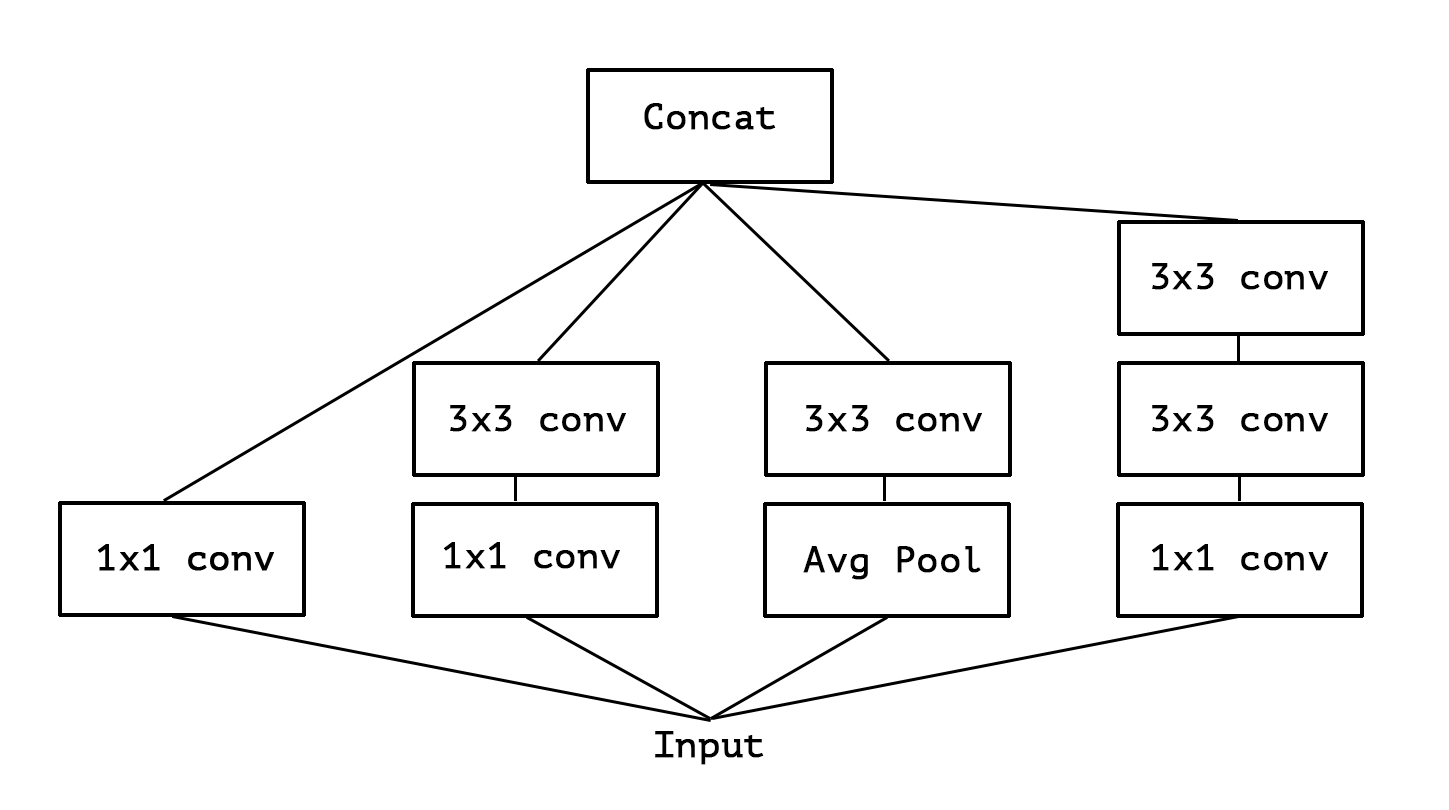}
\end{figure}

This idea behind the Inception module is to make this process easier and more efficient by explicitly factoring it into a series of operations that would independently look at cross-channel correlations and at spatial correlations. More precisely, the typical Inception module first looks at cross-channel correlations via a set of 1x1 convolutions, mapping the input data into 3 or 4 separate spaces that are smaller than the original input space, and then maps all correlations in these smaller 3D spaces, via regular 3x3 or 5x5 convolutions. This is illustrated in figure \ref{inception_module}. In effect, the fundamental hypothesis behind Inception is that cross-channel correlations and spatial correlations are sufficiently decoupled that it is preferable not to map them jointly \footnote{A variant of the process is to independently look at width-wise correlations and height-wise correlations. This is implemented by some of the modules found in Inception V3, which alternate 7x1 and 1x7 convolutions. The use of such spatially separable convolutions has a long history in image processing and has been used in some convolutional neural network implementations since at least 2012 (possibly earlier).}.

\begin{figure}[!ht]
  \caption{A simplified Inception module.}
  \label{simplified_inception_module}
  \centering
    \includegraphics[width=0.45\textwidth]{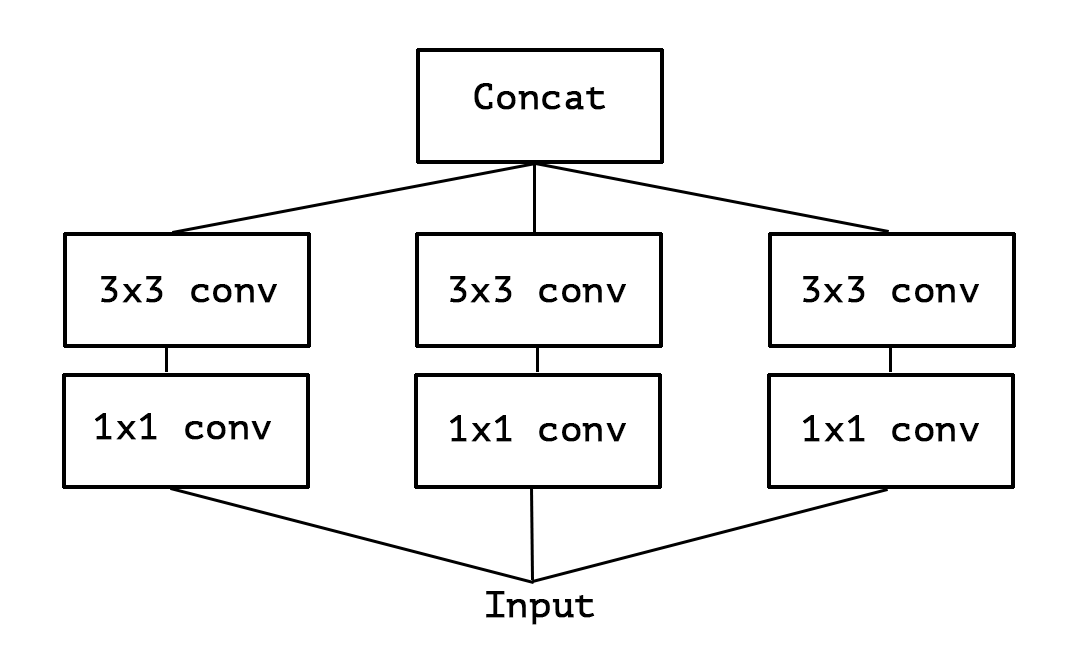}
\end{figure}

\begin{figure}[!ht]
  \caption{A strictly equivalent reformulation of the simplified Inception module.}
  \label{simplified_inception_reformulation}
  \centering
    \includegraphics[width=0.45\textwidth]{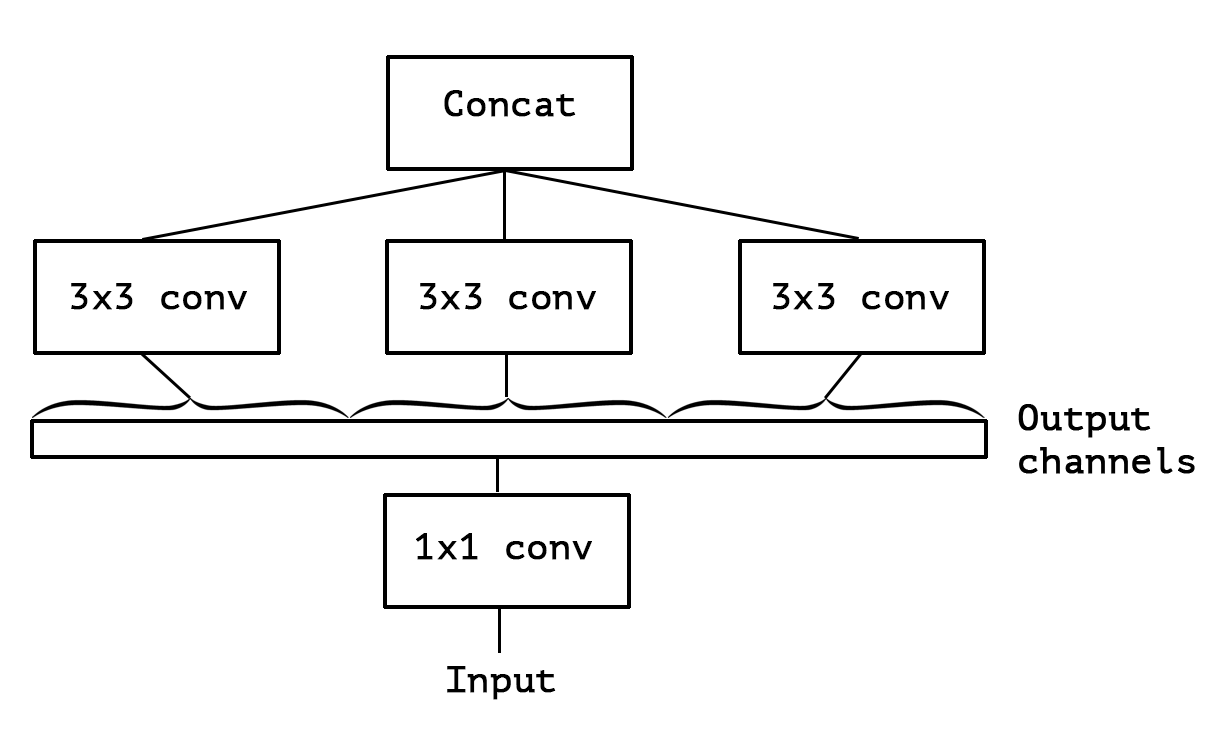}
\end{figure}

Consider a simplified version of an Inception module that only uses one size of convolution (e.g. 3x3) and does not include an average pooling tower (figure \ref{simplified_inception_module}). This Inception module can be reformulated as a large 1x1 convolution followed by spatial convolutions that would operate on non-overlapping segments of the output channels (figure \ref{simplified_inception_reformulation}). This observation naturally raises the question: what is the effect of the number of segments in the partition (and their size)? Would it be reasonable to make a much stronger hypothesis than the Inception hypothesis, and assume that cross-channel correlations and spatial correlations can be mapped completely separately?

\begin{figure}[!ht]
  \caption{An ``extreme'' version of our Inception module, with one spatial convolution per output channel of the 1x1 convolution.}
  \label{extreme_inception}
  \centering
    \includegraphics[width=0.45\textwidth]{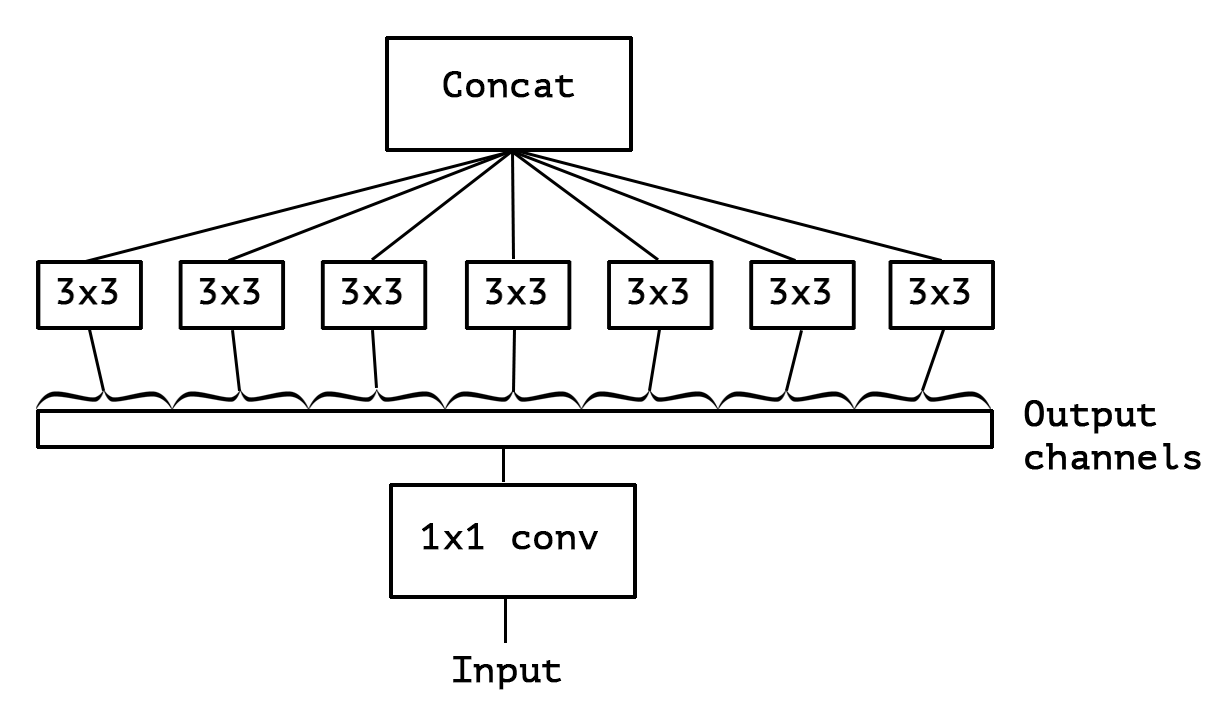}
\end{figure}

\subsection{The continuum between convolutions and separable convolutions} 

An ``extreme'' version of an Inception module, based on this stronger hypothesis, would first use a 1x1 convolution to map cross-channel correlations, and would then separately map the spatial correlations of every output channel. This is shown in figure \ref{extreme_inception}. We remark that this extreme form of an Inception module is almost identical to a \textit{depthwise separable convolution}, an operation that has been used in neural network design as early as 2014 \cite{sifre14} and has become more popular since its inclusion in the TensorFlow framework \cite{tensorflow2015-whitepaper} in 2016.

A depthwise separable convolution, commonly called ``separable convolution'' in deep learning frameworks such as TensorFlow and Keras, consists in a \textit{depthwise convolution}, i.e. a spatial convolution performed independently over each channel of an input, followed by a \textit{pointwise convolution}, i.e. a 1x1 convolution, projecting the channels output by the depthwise convolution onto a new channel space. This is not to be confused with a spatially separable convolution, which is also commonly called ``separable convolution'' in the image processing community.

Two minor differences between and ``extreme'' version of an Inception module and a depthwise separable convolution would be:

\begin{itemize}
  \item The order of the operations: depthwise separable convolutions as usually implemented (e.g. in TensorFlow) perform first channel-wise spatial convolution and then perform 1x1 convolution, whereas Inception performs the 1x1 convolution first.
  \item The presence or absence of a non-linearity after the first operation. In Inception, both operations are followed by a ReLU non-linearity, however depthwise separable convolutions are usually implemented without non-linearities.
\end{itemize}

We argue that the first difference is unimportant, in particular because these operations are meant to be used in a stacked setting. The second difference might matter, and we investigate it in the experimental section (in particular see figure \ref{xception_imagenet_activations}).

We also note that other intermediate formulations of Inception modules that lie in between regular Inception modules and depthwise separable convolutions are also possible: in effect, there is a discrete spectrum between regular convolutions and depthwise separable convolutions, parametrized by the number of independent channel-space segments used for performing spatial convolutions. A regular convolution (preceded by a 1x1 convolution), at one extreme of this spectrum, corresponds to the single-segment case; a depthwise separable convolution corresponds to the other extreme where there is one segment per channel; Inception modules lie in between, dividing a few hundreds of channels into 3 or 4 segments. The properties of such intermediate modules appear not to have been explored yet.

Having made these observations, we suggest that it may be possible to improve upon the Inception family of architectures by replacing Inception modules with depthwise separable convolutions, i.e. by building models that would be stacks of depthwise separable convolutions. This is made practical by the efficient depthwise convolution implementation available in TensorFlow. In what follows, we present a convolutional neural network architecture based on this idea, with a similar number of parameters as Inception V3, and we evaluate its performance against Inception V3 on two large-scale image classification task.

\section{Prior work}

The present work relies heavily on prior efforts in the following areas:

\begin{itemize}
  \item Convolutional neural networks \cite{lecun1995learning, krizhevsky2012imagenet, zeiler2014visualizing}, in particular the VGG-16 architecture \cite{simonyan2014very}, which is schematically similar to our proposed architecture in a few respects.

  \item The Inception architecture family of convolutional neural networks \cite{szegedy2015going, ioffe2015batch, szegedy2015rethinking, inceptionResnet}, which first demonstrated the advantages of factoring convolutions into multiple branches operating successively on channels and then on space.

  \item Depthwise separable convolutions, which our proposed architecture is entirely based upon. While the use of spatially separable convolutions in neural networks has a long history, going back to at least 2012 \cite{mamalet12} (but likely even earlier), the depthwise version is more recent. Laurent Sifre developed depthwise separable convolutions during an internship at Google Brain in 2013, and used them in AlexNet to obtain small gains in accuracy and large gains in convergence speed, as well as a significant reduction in model size. An overview of his work was first made public in a presentation at ICLR 2014 \cite{vanhoucke-iclr14}. Detailed experimental results are reported in Sifre's thesis, section 6.2 \cite{sifre14}. This initial work on depthwise separable convolutions was inspired by prior research from Sifre and Mallat on transformation-invariant scattering \cite{sifre2013, sifre14}. Later, a depthwise separable convolution was used as the first layer of Inception V1 and Inception V2 \cite{szegedy2015going, ioffe2015batch}. Within Google, Andrew Howard \cite{Howard16} has introduced efficient mobile models called MobileNets using depthwise separable convolutions. Jin et al. in 2014 \cite{JinDC14} and Wang et al. in 2016 \cite{WangLF16b} also did related work aiming at reducing the size and computational cost of convolutional neural networks using separable convolutions. Additionally, our work is only possible due to the inclusion of an efficient implementation of depthwise separable convolutions in the TensorFlow framework \cite{tensorflow2015-whitepaper}.

  \item Residual connections, introduced by He et al. in \cite{he2015deep}, which our proposed architecture uses extensively.
\end{itemize}

\section{The Xception architecture}

We propose a convolutional neural network architecture based entirely on depthwise separable convolution layers. In effect, we make the following hypothesis: that the mapping of cross-channels correlations and spatial correlations in the feature maps of convolutional neural networks can be \textit{entirely} decoupled. Because this hypothesis is a stronger version of the hypothesis underlying the Inception architecture, we name our proposed architecture \textit{Xception}, which stands for ``Extreme Inception''.

\begin{figure*}
  \caption{The Xception architecture: the data first goes through the entry flow, then through the middle flow which is repeated eight times, and finally through the exit flow. Note that all Convolution and SeparableConvolution layers are followed by batch normalization \cite{ioffe2015batch} (not included in the diagram). All SeparableConvolution layers use a depth multiplier of 1 (no depth expansion).}
  \label{xception_architecture_final}
  \centering
    \includegraphics[width=1\textwidth]{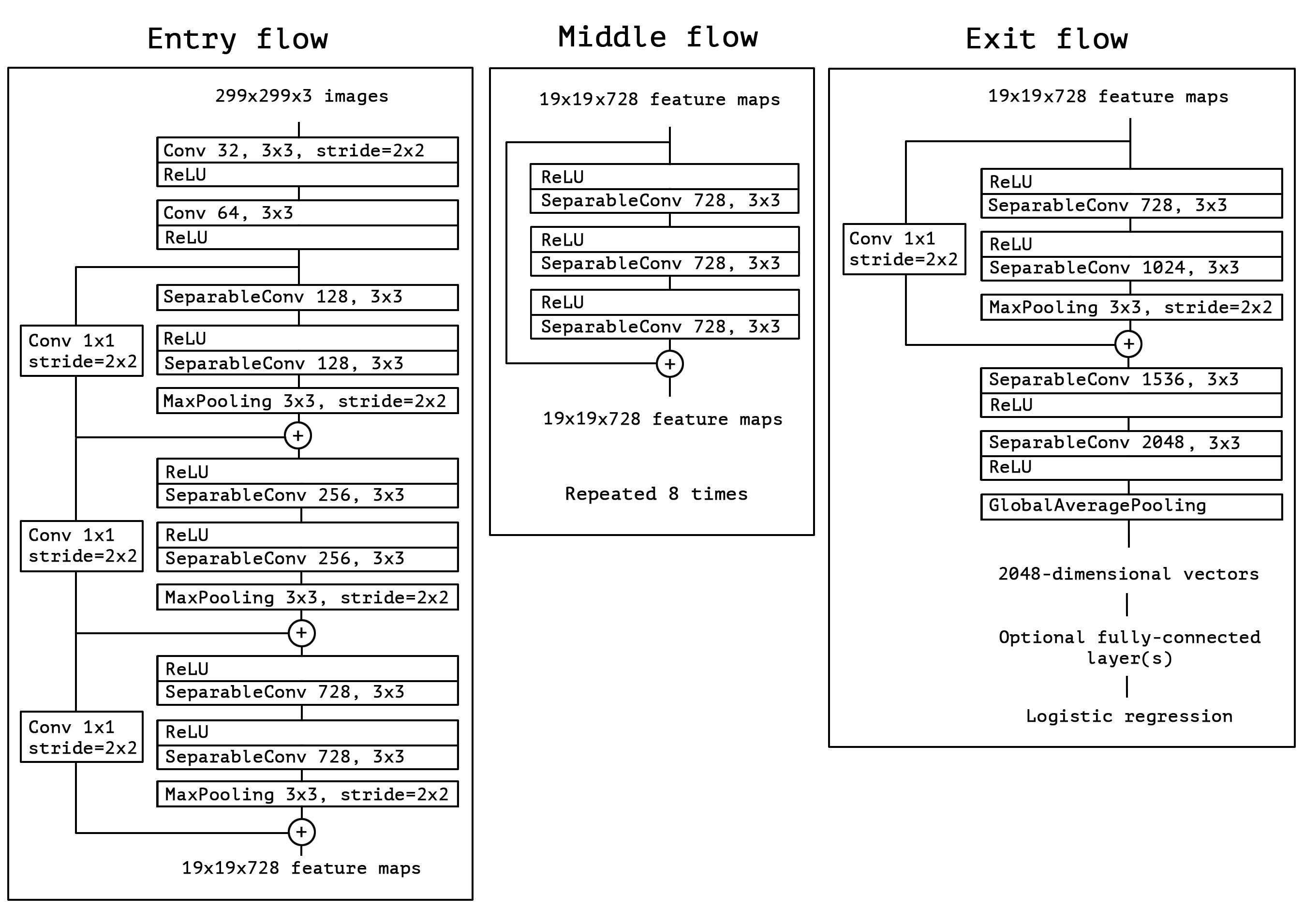}
\end{figure*}

A complete description of the specifications of the network is given in figure \ref{xception_architecture_final}. The Xception architecture has 36 convolutional layers forming the feature extraction base of the network. In our experimental evaluation we will exclusively investigate image classification and therefore our convolutional base will be followed by a logistic regression layer. Optionally one may insert fully-connected layers before the logistic regression layer, which is explored in the experimental evaluation section (in particular, see figures \ref{xception_jft_no_fc} and \ref{xception_jft_with_fc}). The 36 convolutional layers are structured into 14 modules, all of which have linear residual connections around them, except for the first and last modules.

In short, the Xception architecture is a linear stack of depthwise separable convolution layers with residual connections. This makes the architecture very easy to define and modify; it takes only 30 to 40 lines of code using a high-level library such as Keras \cite{chollet2015keras} or TensorFlow-Slim \cite{tfslim}, not unlike an architecture such as VGG-16 \cite{simonyan2014very}, but rather unlike architectures such as Inception V2 or V3 which are far more complex to define. An open-source implementation of Xception using Keras and TensorFlow is provided as part of the Keras Applications module\footnote{\url{https://keras.io/applications/\#xception}}, under the MIT license.

\section{Experimental evaluation}

We choose to compare Xception to the Inception V3 architecture, due to their similarity of scale: Xception and Inception V3 have nearly the same number of parameters (table \ref{sizeandspeed}), and thus any performance gap could not be attributed to a difference in network capacity. We conduct our comparison on two image classification tasks: one is the well-known 1000-class single-label classification task on the ImageNet dataset \cite{russakovsky2014imagenet}, and the other is a 17,000-class multi-label classification task on the large-scale JFT dataset.

\subsection{The JFT dataset}

JFT is an internal Google dataset for large-scale image classification dataset, first introduced by Hinton et al. in \cite{hinton15}, which comprises over 350 million high-resolution images annotated with labels from a set of 17,000 classes. To evaluate the performance of a model trained on JFT, we use an auxiliary dataset, \textbf{FastEval14k}.

FastEval14k is a dataset of 14,000 images with dense annotations from about 6,000 classes (36.5 labels per image on average). On this dataset we evaluate performance using Mean Average Precision for top 100 predictions (MAP@100), and we weight the contribution of each class to MAP@100 with a score estimating how common (and therefore important) the class is among social media images. This evaluation procedure is meant to capture performance on frequently occurring labels from social media, which is crucial for production models at Google.

\subsection{Optimization configuration}

A different optimization configuration was used for ImageNet and JFT:

\begin{itemize}
\item On ImageNet:
  \begin{itemize}
  \item Optimizer: SGD
  \item Momentum: 0.9
  \item Initial learning rate: 0.045
  \item Learning rate decay: decay of rate 0.94 every 2 epochs
  \end{itemize}

\item On JFT:
  \begin{itemize}
  \item Optimizer: RMSprop \cite{rmsprop}
  \item Momentum: 0.9
  \item Initial learning rate: 0.001
  \item Learning rate decay: decay of rate 0.9 every 3,000,000 samples
  \end{itemize}

\end{itemize}

For both datasets, the same exact same optimization configuration was used for both Xception and Inception V3. Note that this configuration was tuned for best performance with Inception V3; we did not attempt to tune optimization hyperparameters for Xception. Since the networks have different training profiles (figure \ref{xception_imagenet}), this may be suboptimal, especially on the ImageNet dataset, on which the optimization configuration used had been carefully tuned for Inception V3.

Additionally, all models were evaluated using Polyak averaging \cite{polyak1992} at inference time.

\subsection{Regularization configuration}

\begin{itemize}
\item \textbf{Weight decay:} The Inception V3 model uses a weight decay (L2 regularization) rate of $4e-5$, which has been carefully tuned for performance on ImageNet. We found this rate to be quite suboptimal for Xception and instead settled for $1e-5$. We did not perform an extensive search for the optimal weight decay rate. The same weight decay rates were used both for the ImageNet experiments and the JFT experiments.

\item \textbf{Dropout:} For the ImageNet experiments, both models include a dropout layer of rate 0.5 before the logistic regression layer. For the JFT experiments, no dropout was included due to the large size of the dataset which made overfitting unlikely in any reasonable amount of time.

\item \textbf{Auxiliary loss tower:} The Inception V3 architecture may optionally include an auxiliary tower which backpropagates the classification loss earlier in the network, serving as an additional regularization mechanism. For simplicity, we choose not to include this auxiliary tower in any of our models.

\end{itemize}

\subsection{Training infrastructure}

All networks were implemented using the TensorFlow framework \cite{tensorflow2015-whitepaper} and trained on 60 NVIDIA K80 GPUs each. For the ImageNet experiments, we used data parallelism with \textit{synchronous} gradient descent to achieve the best classification performance, while for JFT we used \textit{asynchronous} gradient descent so as to speed up training. The ImageNet experiments took approximately 3 days each, while the JFT experiments took over one month each. The JFT models were not trained to full convergence, which would have taken over three month per experiment.

\subsection{Comparison with Inception V3}

\subsubsection{Classification performance}

\begin{figure}[!ht]
  \caption{Training profile on ImageNet}
  \label{xception_imagenet}
  \centering
    \includegraphics[width=0.45\textwidth]{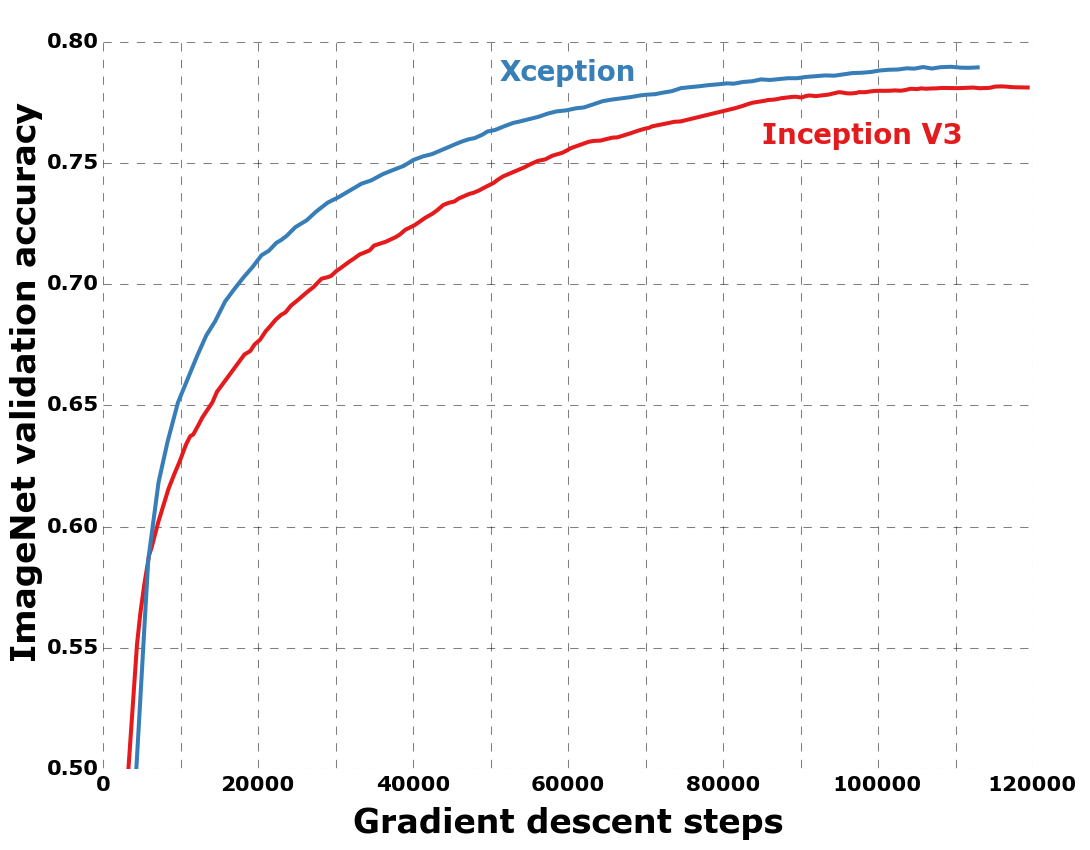}
\end{figure}

\begin{figure}[!ht]
  \caption{Training profile on JFT, without fully-connected layers}
  \label{xception_jft_no_fc}
  \centering
    \includegraphics[width=0.45\textwidth]{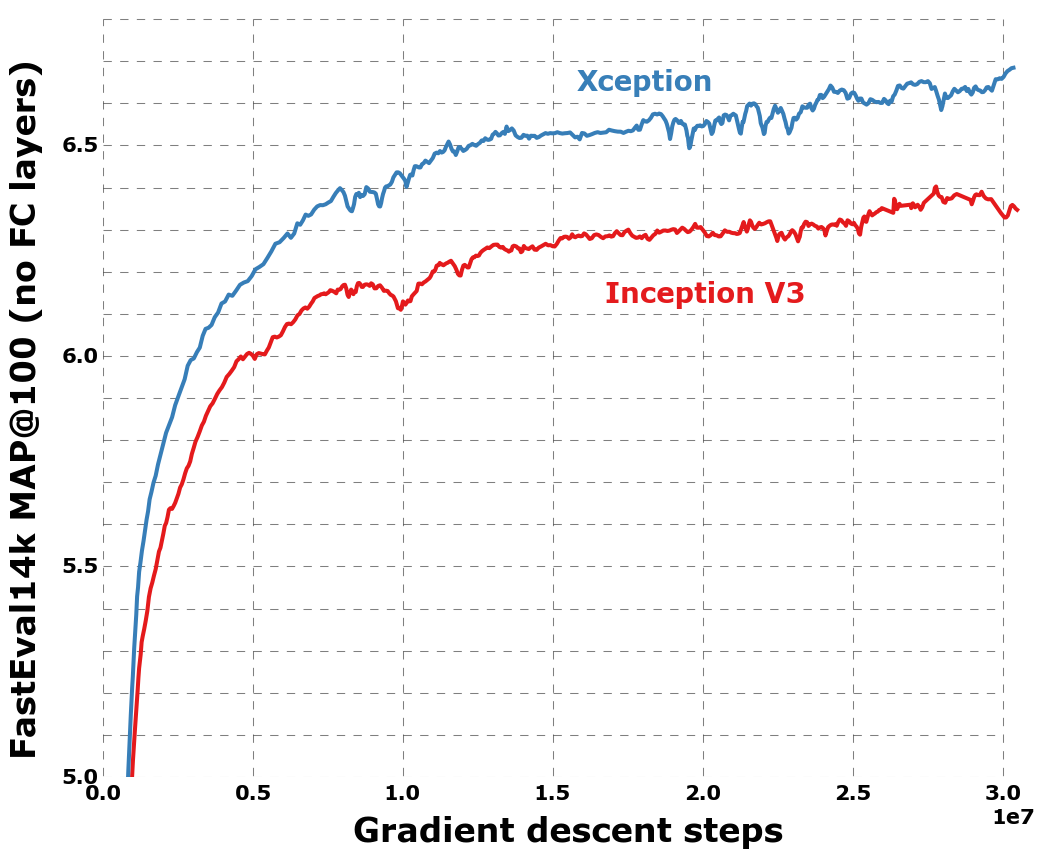}
\end{figure}

\begin{figure}[!ht]
  \caption{Training profile on JFT, with fully-connected layers}
  \label{xception_jft_with_fc}
  \centering
    \includegraphics[width=0.45\textwidth]{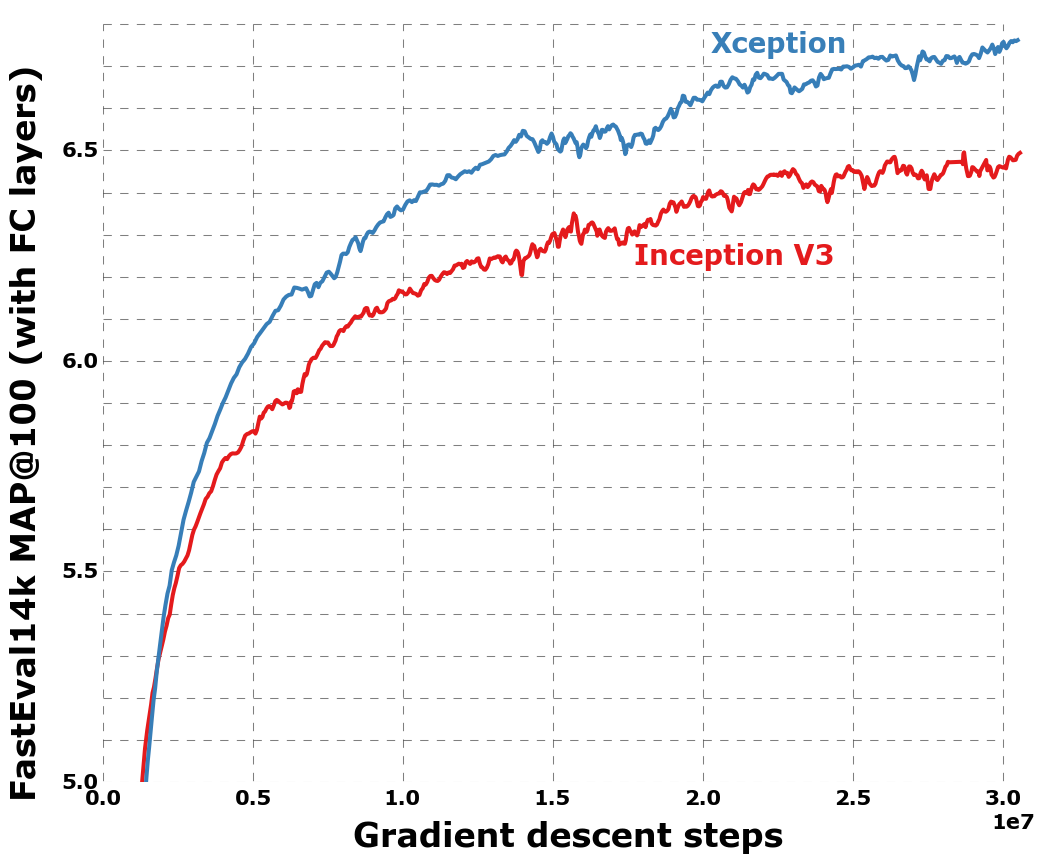}
\end{figure}

All evaluations were run with a single crop of the inputs images and a single model. ImageNet results are reported on the validation set rather than the test set (i.e. on the non-blacklisted images from the validation set of ILSVRC 2012). JFT results are reported after 30 million iterations (one month of training) rather than after full convergence. Results are provided in table \ref{imagenetperf} and table \ref{jftperf}, as well as figure \ref{xception_imagenet}, figure \ref{xception_jft_no_fc}, figure \ref{xception_jft_with_fc}. On JFT, we tested both versions of our networks that did not include any fully-connected layers, and versions that included two fully-connected layers of 4096 units each before the logistic regression layer.

On ImageNet, Xception shows marginally better results than Inception V3. On JFT, Xception shows a ~4.3\% relative improvement on the FastEval14k MAP@100 metric. We also note that Xception outperforms ImageNet results reported by He et al. for ResNet-50, ResNet-101 and ResNet-152 \cite{he2015deep}.

\begin{table}[!ht]
\centering
\caption{Classification performance comparison on ImageNet (single crop, single model). VGG-16 and ResNet-152 numbers are only included as a reminder. The version of Inception V3 being benchmarked does not include the auxiliary tower.}
\label{imagenetperf}

  \begin{tabular}{lcc}
  \toprule
              & \textbf{Top-1 accuracy}    & \textbf{Top-5 accuracy} \\ \hline
  \textbf{VGG-16}        &   0.715        &     0.901              \\ \hline
  \textbf{ResNet-152}    &   0.770        &     0.933              \\ \hline
  \textbf{Inception V3}  &   0.782        &     0.941              \\ \hline
  \textbf{Xception}      & \textbf{0.790} & \textbf{0.945}         \\ \hline

  \end{tabular}

\end{table}

\begin{table}[!ht]
\centering
\caption{Classification performance comparison on JFT (single crop, single model).}
\label{jftperf}

  \begin{tabular}{lc}
  \toprule
                         & \textbf{FastEval14k MAP@100}   \\ \hline
  \textbf{Inception V3 - no FC layers}    &   6.36            \\ \hline
  \textbf{Xception - no FC layers}        &   6.70            \\ \hline
  \textbf{Inception V3 with FC layers}  &   6.50            \\ \hline
  \textbf{Xception with FC layers}      & \textbf{6.78}     \\ \hline

  \end{tabular}

\end{table}

The Xception architecture shows a much larger performance improvement on the JFT dataset compared to the ImageNet dataset. We believe this may be due to the fact that Inception V3 was developed with a focus on ImageNet and may thus be by design over-fit to this specific task. On the other hand, neither architecture was tuned for JFT. It is likely that a search for better hyperparameters for Xception on ImageNet (in particular optimization parameters and regularization parameters) would yield significant additional improvement.

\subsubsection{Size and speed}

\begin{table}[!ht]
\centering
\caption{Size and training speed comparison.}
\label{sizeandspeed}

  \begin{tabular}{lcc}
  \toprule
               & \textbf{Parameter count} & \textbf{Steps/second} \\ \hline
  \textbf{Inception V3}  &   23,626,728   &     31                 \\ \hline
  \textbf{Xception} &       22,855,952   &     28                   \\ \hline

  \end{tabular}

\end{table}

In table \ref{sizeandspeed} we compare the size and speed of Inception V3 and Xception. Parameter count is reported on ImageNet (1000 classes, no fully-connected layers) and the number of training steps (gradient updates) per second is reported on ImageNet with 60 K80 GPUs running synchronous gradient descent. Both architectures have approximately the same size (within 3.5\%), and Xception is marginally slower. We expect that engineering optimizations at the level of the depthwise convolution operations can make Xception faster than Inception V3 in the near future. The fact that both architectures have almost the same number of parameters indicates that the improvement seen on ImageNet and JFT does not come from added capacity but rather from a more efficient use of the model parameters.

\subsection{Effect of the residual connections}

\begin{figure}[!ht]
  \caption{Training profile with and without residual connections.}
  \label{xception_imagenet_nonresidual}
  \centering
    \includegraphics[width=0.45\textwidth]{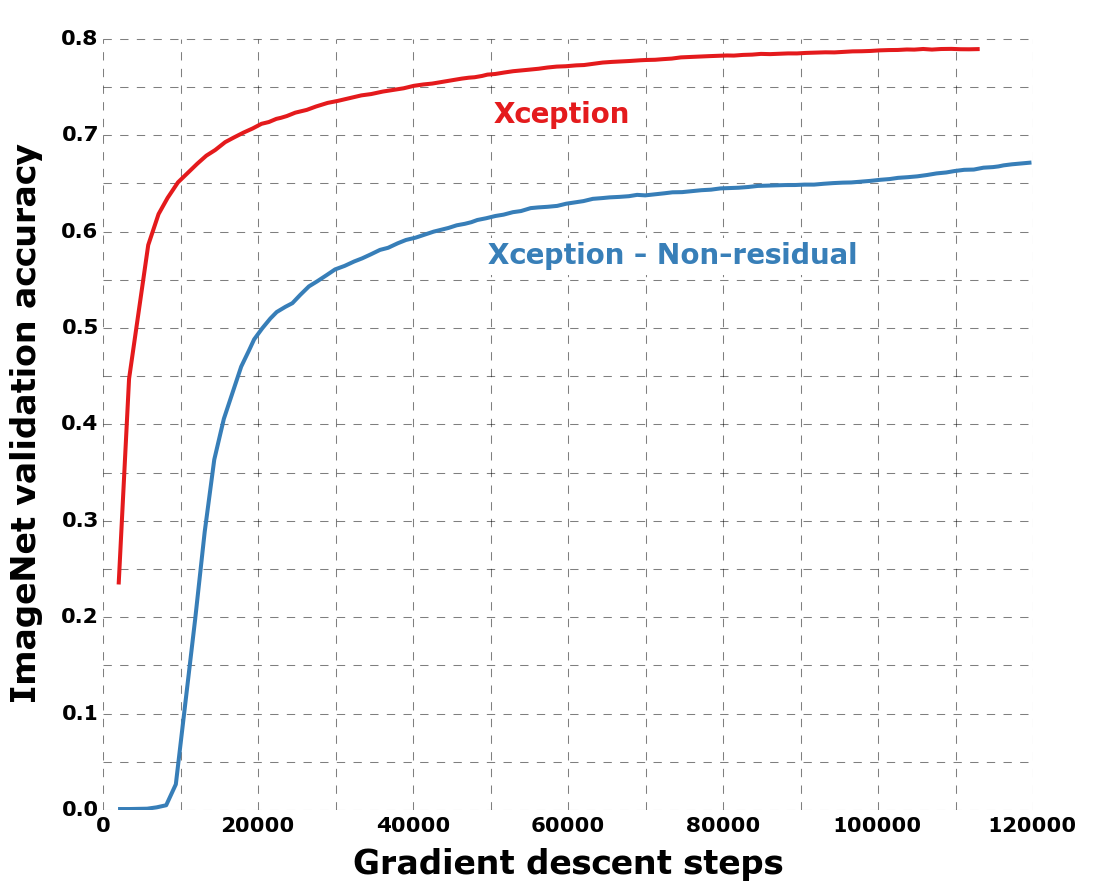}
\end{figure}

To quantify the benefits of residual connections in the Xception architecture, we benchmarked on ImageNet a modified version of Xception that does not include any residual connections. Results are shown in figure \ref{xception_imagenet_nonresidual}. Residual connections are clearly essential in helping with convergence, both in terms of speed and final classification performance. However we will note that benchmarking the non-residual model with the same optimization configuration as the residual model may be uncharitable and that better optimization configurations might yield more competitive results.

Additionally, let us note that this result merely shows the importance of residual connections \textit{for this specific architecture}, and that residual connections are in no way \textit{required} in order to build models that are stacks of depthwise separable convolutions. We also obtained excellent results with non-residual VGG-style models where all convolution layers were replaced with depthwise separable convolutions (with a depth multiplier of 1), superior to Inception V3 on JFT at equal parameter count.

\subsection{Effect of an intermediate activation after pointwise convolutions}

\begin{figure}[!ht]
  \caption{Training profile with different activations between the depthwise and pointwise operations of the separable convolution layers.}
  \label{xception_imagenet_activations}
  \centering
    \includegraphics[width=0.45\textwidth]{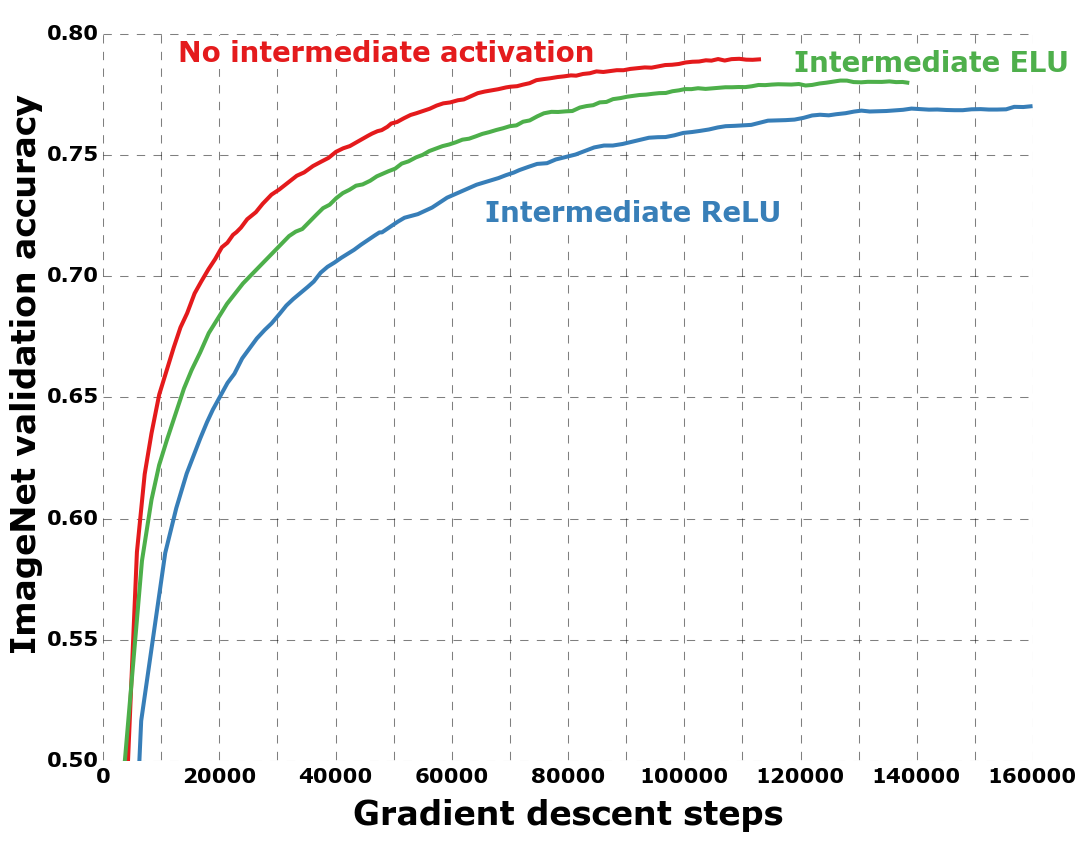}
\end{figure}

We mentioned earlier that the analogy between depthwise separable convolutions and Inception modules suggests that depthwise separable convolutions should potentially include a non-linearity between the depthwise and pointwise operations. In the experiments reported so far, no such non-linearity was included. However we also experimentally tested the inclusion of either ReLU or ELU \cite{clevert15} as intermediate non-linearity. Results are reported on ImageNet in figure \ref{xception_imagenet_activations}, and show that the absence of any non-linearity leads to both faster convergence and better final performance.

This is a remarkable observation, since Szegedy et al. report the opposite result in \cite{szegedy2015rethinking} for Inception modules. It may be that the depth of the intermediate feature spaces on which spatial convolutions are applied is critical to the usefulness of the non-linearity: for deep feature spaces (e.g. those found in Inception modules) the non-linearity is helpful, but for shallow ones (e.g. the 1-channel deep feature spaces of depthwise separable convolutions) it becomes harmful, possibly due to a loss of information.

\section{Future directions}

We noted earlier the existence of a discrete spectrum between regular convolutions and depthwise separable convolutions, parametrized by the number of independent channel-space segments used for performing spatial convolutions. Inception modules are one point on this spectrum. We showed in our empirical evaluation that the extreme formulation of an Inception module, the depthwise separable convolution, may have advantages over regular a regular Inception module. However, there is no reason to believe that depthwise separable convolutions are optimal. It may be that intermediate points on the spectrum, lying between regular Inception modules and depthwise separable convolutions, hold further advantages. This question is left for future investigation.

\section{Conclusions}

We showed how convolutions and depthwise separable convolutions lie at both extremes of a discrete spectrum, with Inception modules being an intermediate point in between. This observation has led to us to propose replacing Inception modules with depthwise separable convolutions in neural computer vision architectures. We presented a novel architecture based on this idea, named Xception, which has a similar parameter count as Inception V3. Compared to Inception V3, Xception shows small gains in classification performance on the ImageNet dataset and large gains on the JFT dataset. We expect depthwise separable convolutions to become a cornerstone of convolutional neural network architecture design in the future, since they offer similar properties as Inception modules, yet are as easy to use as regular convolution layers.

\begin{small}

{\small
\bibliographystyle{ieee}
\bibliography{xception}

\begin{thebibliography}{10}\itemsep=-1pt

\bibitem{tensorflow2015-whitepaper}
M.~Abadi, A.~Agarwal, P.~Barham, E.~Brevdo, Z.~Chen, C.~Citro, G.~S. Corrado,
  A.~Davis, J.~Dean, M.~Devin, S.~Ghemawat, I.~Goodfellow, A.~Harp, G.~Irving,
  M.~Isard, Y.~Jia, R.~Jozefowicz, L.~Kaiser, M.~Kudlur, J.~Levenberg,
  D.~Man\'{e}, R.~Monga, S.~Moore, D.~Murray, C.~Olah, M.~Schuster, J.~Shlens,
  B.~Steiner, I.~Sutskever, K.~Talwar, P.~Tucker, V.~Vanhoucke, V.~Vasudevan,
  F.~Vi\'{e}gas, O.~Vinyals, P.~Warden, M.~Wattenberg, M.~Wicke, Y.~Yu, and
  X.~Zheng.
\newblock {TensorFlow}: Large-scale machine learning on heterogeneous systems,
  2015.
\newblock Software available from tensorflow.org.

\bibitem{chollet2015keras}
F.~Chollet.
\newblock Keras.
\newblock https://github.com/fchollet/keras, 2015.

\bibitem{clevert15}
D.-A. Clevert, T.~Unterthiner, and S.~Hochreiter.
\newblock Fast and accurate deep network learning by exponential linear units
  (elus).
\newblock {\em arXiv preprint arXiv:1511.07289}, 2015.

\bibitem{he2015deep}
K.~He, X.~Zhang, S.~Ren, and J.~Sun.
\newblock Deep residual learning for image recognition.
\newblock {\em arXiv preprint arXiv:1512.03385}, 2015.

\bibitem{hinton15}
G.~Hinton, O.~Vinyals, and J.~Dean.
\newblock Distilling the knowledge in a neural network, 2015.

\bibitem{Howard16}
A.~Howard.
\newblock Mobilenets: Efficient convolutional neural networks for mobile vision
  applications.
\newblock Forthcoming.

\bibitem{ioffe2015batch}
S.~Ioffe and C.~Szegedy.
\newblock Batch normalization: Accelerating deep network training by reducing
  internal covariate shift.
\newblock In {\em Proceedings of The 32nd International Conference on Machine
  Learning}, pages 448--456, 2015.

\bibitem{JinDC14}
J.~Jin, A.~Dundar, and E.~Culurciello.
\newblock Flattened convolutional neural networks for feedforward acceleration.
\newblock {\em arXiv preprint arXiv:1412.5474}, 2014.

\bibitem{krizhevsky2012imagenet}
A.~Krizhevsky, I.~Sutskever, and G.~E. Hinton.
\newblock Imagenet classification with deep convolutional neural networks.
\newblock In {\em Advances in neural information processing systems}, pages
  1097--1105, 2012.

\bibitem{lecun1995learning}
Y.~LeCun, L.~Jackel, L.~Bottou, C.~Cortes, J.~S. Denker, H.~Drucker, I.~Guyon,
  U.~Muller, E.~Sackinger, P.~Simard, et~al.
\newblock Learning algorithms for classification: A comparison on handwritten
  digit recognition.
\newblock {\em Neural networks: the statistical mechanics perspective},
  261:276, 1995.

\bibitem{lin2013network}
M.~Lin, Q.~Chen, and S.~Yan.
\newblock Network in network.
\newblock {\em arXiv preprint arXiv:1312.4400}, 2013.

\bibitem{mamalet12}
F.~{Mamalet} and C.~{Garcia}.
\newblock {Simplifying ConvNets for Fast Learning}.
\newblock In {\em International Conference on Artificial Neural Networks (ICANN
  2012)}, pages 58--65. Springer, 2012.

\bibitem{polyak1992}
B.~T. Polyak and A.~B. Juditsky.
\newblock Acceleration of stochastic approximation by averaging.
\newblock {\em SIAM J. Control Optim.}, 30(4):838--855, July 1992.

\bibitem{russakovsky2014imagenet}
O.~Russakovsky, J.~Deng, H.~Su, J.~Krause, S.~Satheesh, S.~Ma, Z.~Huang,
  A.~Karpathy, A.~Khosla, M.~Bernstein, et~al.
\newblock Imagenet large scale visual recognition challenge.
\newblock 2014.

\bibitem{sifre14}
L.~Sifre.
\newblock Rigid-motion scattering for image classification, 2014.
\newblock Ph.D. thesis.

\bibitem{sifre2013}
L.~Sifre and S.~Mallat.
\newblock Rotation, scaling and deformation invariant scattering for texture
  discrimination.
\newblock In {\em 2013 {IEEE} Conference on Computer Vision and Pattern
  Recognition, Portland, OR, USA, June 23-28, 2013}, pages 1233--1240, 2013.

\bibitem{tfslim}
N.~Silberman and S.~Guadarrama.
\newblock Tf-slim, 2016.

\bibitem{simonyan2014very}
K.~Simonyan and A.~Zisserman.
\newblock Very deep convolutional networks for large-scale image recognition.
\newblock {\em arXiv preprint arXiv:1409.1556}, 2014.

\bibitem{inceptionResnet}
C.~Szegedy, S.~Ioffe, and V.~Vanhoucke.
\newblock Inception-v4, inception-resnet and the impact of residual connections
  on learning.
\newblock {\em arXiv preprint arXiv:1602.07261}, 2016.

\bibitem{szegedy2015going}
C.~Szegedy, W.~Liu, Y.~Jia, P.~Sermanet, S.~Reed, D.~Anguelov, D.~Erhan,
  V.~Vanhoucke, and A.~Rabinovich.
\newblock Going deeper with convolutions.
\newblock In {\em Proceedings of the IEEE Conference on Computer Vision and
  Pattern Recognition}, pages 1--9, 2015.

\bibitem{szegedy2015rethinking}
C.~Szegedy, V.~Vanhoucke, S.~Ioffe, J.~Shlens, and Z.~Wojna.
\newblock Rethinking the inception architecture for computer vision.
\newblock {\em arXiv preprint arXiv:1512.00567}, 2015.

\bibitem{rmsprop}
T.~Tieleman and G.~Hinton.
\newblock Divide the gradient by a running average of its recent magnitude.
\newblock COURSERA: Neural Networks for Machine Learning, 4, 2012.
\newblock Accessed: 2015-11-05.

\bibitem{vanhoucke-iclr14}
V.~Vanhoucke.
\newblock Learning visual representations at scale.
\newblock ICLR, 2014.

\bibitem{WangLF16b}
M.~Wang, B.~Liu, and H.~Foroosh.
\newblock Factorized convolutional neural networks.
\newblock {\em arXiv preprint arXiv:1608.04337}, 2016.

\bibitem{zeiler2014visualizing}
M.~D. Zeiler and R.~Fergus.
\newblock Visualizing and understanding convolutional networks.
\newblock In {\em Computer Vision--ECCV 2014}, pages 818--833. Springer, 2014.

\end{thebibliography}
}

\end{small}

\end{document}